# Digital Phenotyping for Adolescent Mental Health: A Feasibility Study Employing Machine Learning to Predict Mental Health Risk From Active and Passive Smartphone Data

**Balasundaram Kadirvelu**[1*], **Teresa Bellido Bel**[2*], **Aglaia Freccero**[2], **Martina Di Simplicio**[2], **Dasha Nicholls**[2#$], **A Aldo Faisal**[1, 3^$]

[1] Brain & Behaviour Lab, Department of Computing and Bioengineering, Imperial, London, UK
[2] Division of Psychiatry, Department of Brain Sciences, Imperial, London, UK
[3] Chair in Digital Health, Universität Bayreuth, Bayreuth, Germany

[*] Joint first authors [$] Joint last authors
[#] Clinical corresponding author e-mail: d.nicholls@imperial.ac.uk
[^] Technical corresponding author e-mail: a.faisal@imperial.ac.uk

## Abstract

**Background:**
Adolescents are particularly vulnerable to mental disorders, with over 75% of cases manifesting before the age of 25. Research indicates that only 18 to 34% of young people experiencing high levels of depression or anxiety symptoms seek support. Digital tools leveraging smartphones offer scalable and cost-effective early mental health intervention opportunities. Active (self-reported) and passive (sensor-based) data collected through smartphones enable digital phenotyping, providing rich insights into behavioural and environmental factors influencing mental health. Despite these advances, integrating these data streams from non-clinical adolescent populations remains underexplored.

**Objective:**
This study aimed to evaluate the feasibility of integrating active and passive smartphone data to predict mental health outcomes in non-clinical adolescents using a novel machine learning framework. Specifically, we investigated the utility of the Mindcraft app in predicting risks for internalising and externalising disorders, eating disorders, insomnia and suicidal ideation, with an emphasis on improving prediction accuracy through data integration and advanced modelling techniques.

**Methods:**
Participants (N=103; mean age 16.1 years) were recruited from three London schools. At baseline, participants completed the Strengths and Difficulties Questionnaire (SDQ), the Eating Disorders-15 Questionnaire (ED-15), the Sleep Condition Indicator Questionnaire (SCI) and indicated the presence/absence of suicidal ideation. They used the Mindcraft app for 14 days, contributing active data via self-reports such as mood, sleep, and loneliness and passive data from smartphone sensors such as step count, location, and ambient noise. A contrastive pretraining phase was applied to enhance user-specific feature stability, followed by supervised fine-tuning. The model evaluation employed leave-one-subject-out cross-validation using balanced accuracy as the primary metric. Comparative analyses were conducted with CatBoost and MLP models without pretraining. SHAP values provided interpretability for feature contributions.

**Results:**
The integration of active and passive data achieved superior performance compared to individual data sources, with mean balanced accuracies of 0.71 for SDQ-High risk, 0.67 for insomnia, 0.77 for suicidal ideation and 0.70 for eating disorders. The contrastive learning framework stabilised daily behavioural representations, enhancing predictive robustness. SHAP analysis revealed clinically meaningful features, such as negative thinking and location entropy, underscoring the complementary nature of active and passive data.

**Conclusions:**
This study demonstrates the potential of integrating active and passive smartphone data with advanced machine learning techniques for predicting adolescent mental health risks. By using innovative machine learning approaches, such as contrastive learning, and leveraging a scalable platform like Mindcraft, we establish a comprehensive framework for identifying early mental health challenges across a range of outcomes. These results pave the way for developing more accessible strategies to support early detection and interventions in adolescent mental health.

## Introduction

Children and young people (CYP) are particularly vulnerable to mental health problems due to critical developmental changes in emotion, behaviour, and cognition [1], with over 75% of such disorders emerging before the age of 25 [2]. The global prevalence of mental disorders in CYP is estimated at 13.4%, with anxiety, depression, and eating disorders among the most prevalent and sleep problems being a main risk factor and symptom [3, 4]. Research indicates that only 18 to 34% of young people experiencing high levels of depression or anxiety symptoms seek professional support[5]. This critical gap in receiving care underscores an urgent need for scalable, accessible, and youth-friendly solutions in mental health care.

Internalising and externalising are two broad categories of emotional and behavioural problems, which in adolescence are associated with an increased likelihood of developing a psychiatric disorder later in adulthood [6]. While internalising problem behaviour is focused on the self (e.g. withdrawal, anxiety, depression, emotional problems) [7], externalising problem behaviour particularly occurs in interaction with the social environment (e.g. aggression, impulsivity, deviance, hyperactivity) [8]. School-based research suggests that the school environment plays a critical role in shaping students' behaviour and has a significant impact on CYP's health [9]. This makes schools well-placed to identify and address multiple determinants of mental health risk at the individual and community levels [10].

Digital health interventions offer a promising, cost-effective, and scalable approach to mental health prevention and early intervention in CYP. Recent years have seen rapid growth in mental health applications [11-13], which leverage smartphones' unique data collection capabilities. Emerging evidence indicates that smartphone-measured lifestyle factors correlate closely with mental health outcomes [14, 15]. Smartphones capture active data (subjective self-reports on behaviours and experiences) and passive data (objective sensor-based information such as GPS, accelerometer, and microphone usage), providing behavioural markers—collectively termed "digital phenotyping"—that reveal dynamic interactions between individuals and their environments [16-18].

Integrating active and passive data enables a holistic view of an individual's mental health, blending subjective experiences with objective behavioural markers. By capitalising on smartphones' advanced data collection capabilities, this approach facilitates a real-time capture of mental health conditions, potentially improving the accuracy and timeliness of risk detection [19, 20]. While studies [15, 21-31] have explored active and passive data for mental health monitoring, few have effectively integrated both data types to predict clinically relevant mental health outcomes in CYP. Most focus narrowly on single outcomes, such as depression or anxiety, and are predominantly conducted in clinical populations [22, 30-33], limiting their generalisability to non-clinical adolescent populations.

Machine learning (ML) has shown considerable potential to analyse the complex, multidimensional data generated through digital phenotyping for understanding and predicting mental health states [34, 35]. Preliminary findings suggest that ML-based approaches can support the early detection of adolescent mental health issues [22, 30], highlighting the potential for effective screening tools. However, limited attention has been given to applying ML methods for integrating active and passive data in non-clinical populations of CYP.

This study addresses these gaps by investigating the feasibility of using machine learning to combine active and passive data collected through the Mindcraft app [36] to predict mental health risks in a non-clinical population of CYP. By leveraging both subjective and objective data types and using a novel ML technique, we aim to advance the development of digital screening tools for the early detection of mental health challenges. By focusing on four diverse mental health outcomes, this study explores the potential of integrated digital phenotyping and machine learning to transform adolescent mental health care through early, scalable, and data-driven risk detection.

## Methods

### Recruitment and Data Collection

Participants were recruited from secondary schools in northwest London. We contacted schools via email and followed up via phone. Three schools that expressed interest in taking part in our study were recruited. The inclusion criteria were young people aged 14–18 years, attending years 10 to 13, who had a sufficient level of English to respond to the study instrument and use the app, and who had access to an iOS- or Android-compatible smartphone. We asked for digital informed consent, and parental consent was required for students under 16 years old.

Students initially completed an online survey accessed via a Qualtrics link included in the promotional materials. This survey began with the Strengths and Difficulties Questionnaire (SDQ) [37], a screening tool whose predictions have been largely consistent with clinical diagnoses with good levels of internal consistency and test-retest stability. To detect eating disorders, we included the Eating Disorders-15 Questionnaire (ED-15), which has been described as a valuable tool to quickly assess eating disorder psychopathology in young individuals [38]. We excluded the compensatory behaviours section to simplify the data collection process. Its ability to detect changes early in treatment means that it could be used as a routine outcome measure within therapeutic contexts. We also incorporated a question from the Patient Health Questionnaire version 9 (PHQ-9) [39], which is validated for young people, to identify self-harm/suicidal ideation [40]. Finally, we used the Sleep Condition Indicator (SCI), a brief scale to evaluate insomnia disorder in everyday clinical practice [41, 42].

Upon completing the online survey, participants received a link to download the Mindcraft app [36] from the App Store or Play Store, along with a unique login. Participants were asked to use the app for at least two weeks. The Mindcraft app is a user-friendly mobile application designed to collect self-reported well-being updates (active data) with phone sensor data (passive data). Participants set their data-sharing preferences during onboarding and can adjust them at any time through the app's settings. Detailed technical specifications of the Mindcraft app are available in reference [36].

### Active and Passive Data Features

Once participants began using the app, we gathered active data and eight categories of raw passive data sourced from phone sensors and usage metrics. Active data responses (e.g., mood, sleep quality, and loneliness), scored on a scale of 1 to 7, were directly incorporated as features for the machine learning model. From the passive data, we engineered 92 distinct features (Table 1). To reduce day-to-day noise and enhance the stability of daily feature measurements, we computed the median of each feature across all data points up to the current day for each participant. This cumulative median provided progressively aggregated daily feature values, capturing longitudinal trends while mitigating potential biases from outlier behaviours. This approach ensured stable and consistent input data for machine learning models.

**Table 1 – Summary of features engineered from passive data sensors**

| Passive sensor | Number of features | Feature list |
|---|---|---|
| Ambient light [Android only] | 8 | Total/Mean/Median/Standard deviation of ambient light reading in the day Total/Mean/Median/Standard deviation of ambient light reading during night hours |
| App usage [Android only] | 36 | Total app usage count, Unique apps, Total/Mean/Median time usage in the day Total app usage count, Unique apps, Total/Mean/Median time usage during the night hours Total time in app category of camera/communication/entertainment/gaming/physical health/mental health/Mindcraft/news/productivity/social media Percentage time in app categories of camera/communication/entertainment/gaming/physical health/mental health/Mindcraft/news/productivity/social media |
| Background noise level [Android only] | 10 | Total/Median/Mean /Max/Standard deviation of background noise levels in the day, Total/Median/Mean /Max/Standard deviation of background noise levels during night hours |
| Battery | 8 | Min/Max/Mean/Median of battery level, Num charges per day, Mean battery use per hour, Time below 20 percent, Night time usage count |
| Location | 15 | Mean latitude, Mean longitude, Total distance travelled in a day, Location count, Max distance from home, Mean distance from home, Median distance from home, Night-time movement, Radius of gyration, Standard deviation of latitude, Standard deviation of longitude, Location entropy, Time spent at home |

| Mindcraft usage | 3 | First hour of use, Last hour of use, Nighttime usage, |
| Screen brightness [iOS only] | 8 | Total/Mean/Median/Standard deviation of screen brightness sensor reading in the day<br>Total/Mean/Median/Standard deviation of screen brightness sensor reading during night hours |
| Step count | 4 | Daily step count<br>Is daily step count greater than 5K steps?<br>Is daily step count greater than 7K steps?<br>Is daily step count greater than 10K steps? |

Using the engineered features, we developed a machine-learning model for each of the four mental health outcomes—SDQ risk, insomnia, suicidal ideation, and eating disorders. We employed three distinct feature sets: active data, passive data, and a combination of both. This design enabled us to assess the predictive strength of each feature set individually and in combination, offering insights into their contributions to mental health outcome predictions.

Participants were classified as high-risk or low-risk for each outcome using validated thresholds specific to each mental health measure, framing the prediction task as a binary classification problem. High-risk classifications were defined as follows: for SDQ, a self-reported score of ≥16 [43]; for insomnia, if their Sleep Condition Indicator (SCI) score was ≤16 [42]; for suicidal ideation, if they responded at least once to the question - 'Over the last two weeks, how often have you been bothered by thoughts that you would be better off dead or of hurting yourself in some way? ', and for eating disorders, an ED-15 total score exceeded 2.69, which corresponds to the mean plus one standard deviation in a non-clinical population [44]. The proportions of participants classified as high-risk for each mental health outcome are summarised in Table 1.

### Machine Learning Workflow and Model Development

Figure 1A outlines our machine-learning pipeline, starting with active and passive data collection via the Mindcraft app. The data were pre-processed and engineered to create a comprehensive feature set, which was subjected to a pretraining phase with contrastive learning using triplet margin loss. This pretraining step clustered user-specific features from different days, minimising day-to-day variability and preserving individual behavioural patterns. The resulting stable embeddings were fine-tuned on labelled data in a supervised setting to predict mental health outcomes. This end-to-end pipeline, combining pretraining and fine-tuning, enabled the development of a predictive model evaluated using balanced accuracy and additional metrics.

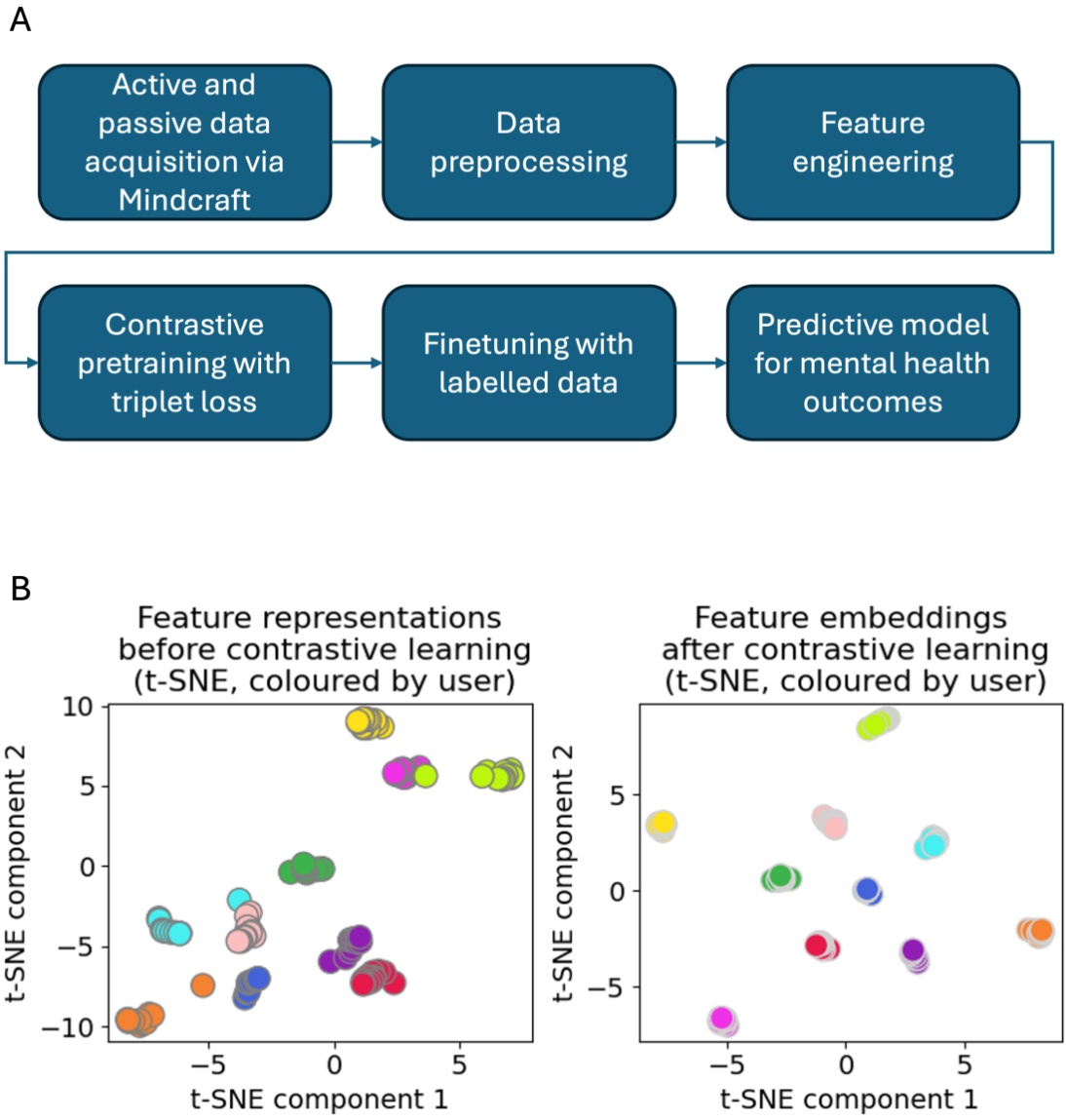

**Figure 1:** (A) Workflow of the machine learning pipeline, from data acquisition to mental health outcome prediction, incorporating contrastive pretraining with triplet loss and fine-tuning. (B) t-SNE visualization of feature embeddings for a sample of 10 test users before (left) and after (right) contrastive pretraining, showing enhanced user-specific clustering following pretraining.

Given the complexity of the data and the potential day-to-day variability in user features, we designed a modelling approach incorporating an innovative pre-training phase with contrastive learning to enhance the stability of user representations across daily measurements. In this phase, triplet margin loss was applied with an anchor data point from a user on one day, a positive data point from the same user on a different day, and a negative data point from a different user.

This approach encouraged the model to cluster user-specific features closely in the latent space, reducing day-to-day variability while preserving user-specific distinctions. The pretraining layer in the architecture uses a neural network with a two-layer MLP to generate embeddings and a projection head consisting of another two-layer MLP for dimensional transformation and effective triplet loss calculations. This adaptation of contrastive learning allowed the model to capture robust and meaningful representations of user characteristics—an essential feature for mental health prediction, where intra-subject variability can hinder accurate classification.

Following pre-training, the model was fine-tuned on labelled data in a supervised phase, using the user-clustered embeddings as input to a two-layer MLP classifier to predict specific mental health outcomes. To address class imbalance, a weighted random sampler and binary cross-entropy loss with a class-weight parameter were employed during training, enhancing the model's ability to distinguish minority classes.

To demonstrate the effect of contrastive pretraining, Figure 1B shows t-SNE visualisations of feature embeddings for a sample of 10 test users. Before pretraining (left plot), user-specific data points were scattered with minimal clustering, reflecting high day-to-day variability. After applying contrastive learning with triplet loss (right plot), data points from the same user formed tighter clusters, indicating enhanced user-specific feature stability.

**Evaluation and Benchmarking**

Day-wise prediction probabilities were averaged for each test user to obtain a single, user-level prediction. Model performance was evaluated using balanced accuracy as the primary metric, along with other relevant metrics such as AUC and F1 score, to assess classification outcomes comprehensively. For interpretability, SHAP values [45] were computed for test folds using DeepExplainer [46], offering insight into the contributions of specific features to classification outcomes.

We validated the model's performance using leave-one-subject-out cross-validation (LOSO CV), where data from all but one user were used for training and validation, with the excluded user's data serving as the test set. This approach ensures the model's generalisability to new individuals, closely simulating real-world applications where accurate predictions for unseen users are critical.

To benchmark our model, we compared it with a CatBoost classifier [47] and a multi-layer perceptron (MLP) network that had a similar number of parameters but was trained without pretraining. Both benchmarks employed class-weight balancing to address the class imbalance in the dataset. CatBoost was selected for its strong performance on tabular data and its built-in capability to handle class imbalance [48], making it well-suited for datasets like ours. These comparisons isolated the effect of contrastive pretraining on model performance.

## Results

### Recruitment and App Usage

Figure 2A provides a conceptual overview of the study, demonstrating how active data (e.g., sleep quality, mood, loneliness) and passive data (e.g., location, app usage, noise levels) collected via the Mindcraft app are integrated into a contrastive learning-based deep neural network to predict mental health outcomes, including SDQ risk, insomnia, suicidal ideation, and eating disorder.

We recruited 103 students from three London schools who downloaded and installed the Mindcraft app. The average age was 16.1 years (SD = 1.0), with 71% identifying as female, 25% as male, and 4% as other/nonbinary. The skew in gender distribution is partially due to one of the participating schools being girls-only. Of the participants, 78 used the app on iPhones, and 25 used Android phones. Table 2 provides demographic information and mental health outcome scores, and Supplementary Figure 1 illustrates the distribution of the different mental health measures across our study population.

**Table 2 –Demographics and mental health measures of the study population**

|  |  |
|---|---|
| N | 103 |
| Female | 73 |
| Age (years) | 16.1±1.0 |
| Strengths and Difficulties Questionnaire score (Mean±SD) | 12.8±6.2 |
| Number (%) scoring in high-risk SDQ category (SDQ score >= 16) | 31 (30.1%) |
| Eating Disorder (EDEQ-15) scale (Mean±SD) | 2.2±1.8 |
| Number (%) scoring in high-risk eating disorder category (EDEQ score >= 2.7) | 38 (36.9%) |
| Sleep Condition Indicator SCI score (Mean±SD) | 19.9±7.8 |
| Number (%) scoring in high-risk insomnia category (SCI score < 17) | 34 (33.0%) |
| Over the last two weeks, how often have you been bothered by thoughts that you would be better off dead or of hurting yourself in some way? (Mean±SD) | 0.6±0.9 |
| Number (%) scoring in high-risk suicidal ideation category (>=1) | 38 (36.9%) |

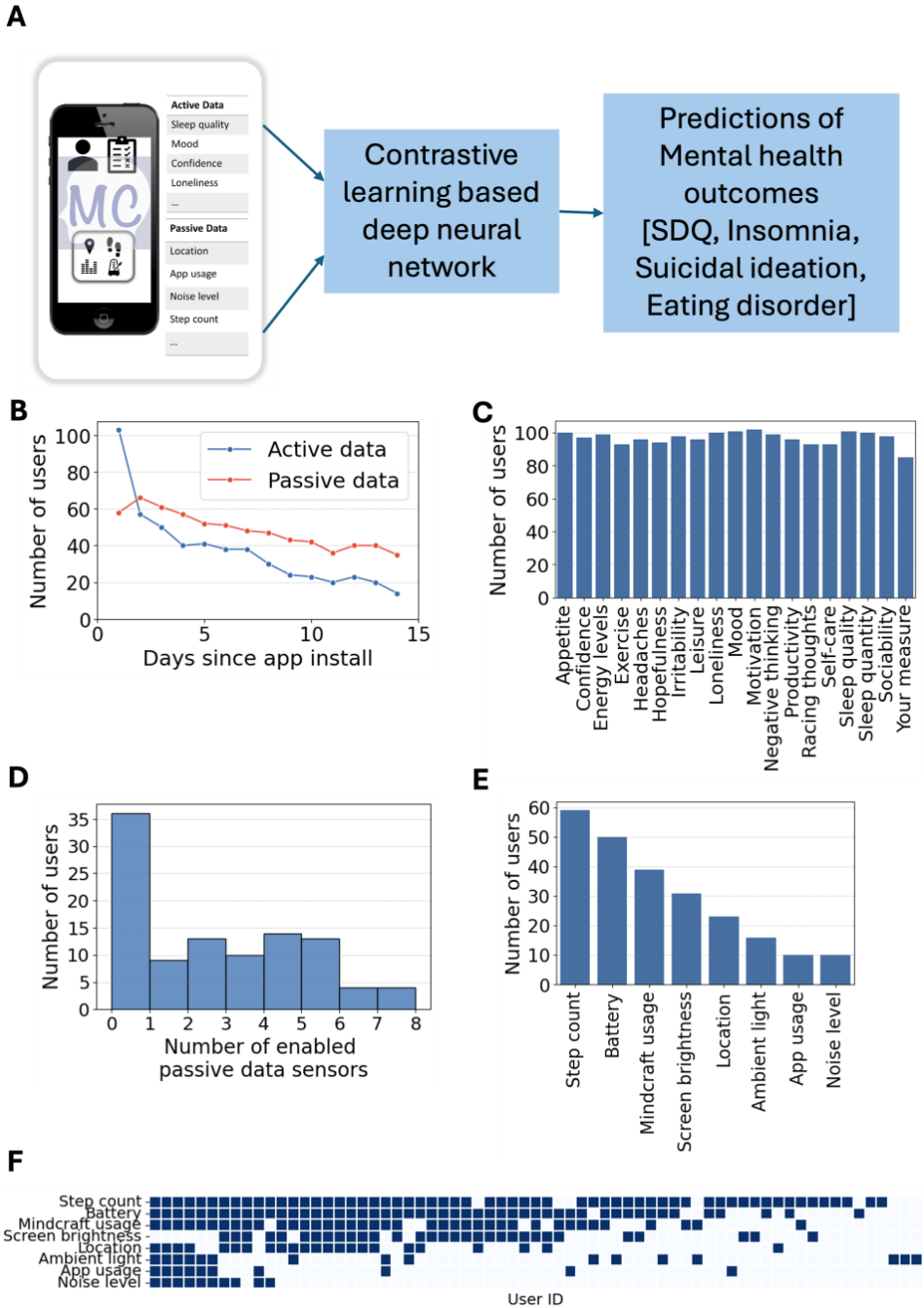

**Figure 2:** (A) Conceptual overview of the study (B) User engagement trends for active and passive data over the 14-day period (C) User participation across active data questions (D) Distribution of enabled passive sensors among users (E) User participation across different passive sensor types (F) Heatmap of passive data completeness by user and sensor type

Participants contributed active data via self-reported measures and passive data through smartphone sensors. Active data included daily ratings of mental well-being measures such as sleep quality, mood, confidence, and loneliness on a 1-7 scale. Passive data comprised data from phone sensors like location, app usage, ambient noise, and step count. Figure 2B shows user engagement patterns over the 14-day study period. Initial engagement was high, with all participants contributing at baseline. However, active data engagement declined more rapidly than passive data, with 14 users contributing active data and 36 users contributing passive data on day 14.

Engagement with active data measures (Figure 2C) remained consistent across users, with slight variations reflecting individual preferences or measure relevance. In contrast, passive data collection exhibited substantial variability (Figure 2D). While 36 users opted not to enable any sensors, others enabled multiple categories. The most frequently enabled sensors were step count and battery usage, followed by Mindcraft usage and screen brightness (Figure 2E). The heatmap visualisation of passive data coverage by users and sensor type (Figure 2F) underscores substantial inter-user variability in passive data coverage, with certain users providing comprehensive data across multiple sensors and others contributing sporadically.

### Exploratory Analysis of Active and Passive Data Features

Figure 3 provides an overview of descriptive statistics and correlations among active and passive data features collected from users. Figure 3A illustrates the distribution of self-reported active data features on a scale of 1 to 7. Positive indicators such as mood, motivation, and confidence had higher mean values than negative indicators such as negative thinking, racing thoughts, and irritability.

The correlation heatmap (Figure 3B) highlights relationships among active data features. The strongest correlation ($r = 0.7$) was observed between motivation and productivity, followed by a strong association between negative thinking and racing thoughts ($r = 0.66$). Positive correlations were also seen among two well-being indicators, such as energy levels and mood ($r = 0.58$), and two distress indicators, like loneliness and negative thinking ($r = 0.57$). Conversely, negative correlations were seen, such as between mood and negative thinking ($r = -0.56$) and irritability and sociability ($r = -0.49$).

Figures 3C-F illustrate four of the 92 engineered passive data features. The distribution of daily step counts (Figure 3C) is right-skewed, with most users taking fewer than 10,000 steps per day. Figure 3D shows the frequency of unique apps opened daily, peaking at 15–20 apps, indicating varying levels of mobile engagement. The entropy of locations visited (Figure 3E) reflects movement variability, with higher values suggesting diverse activity patterns. Finally, Figure 3F

highlights the distribution of mean background noise levels at night, clustering between 30 and 50 dB.

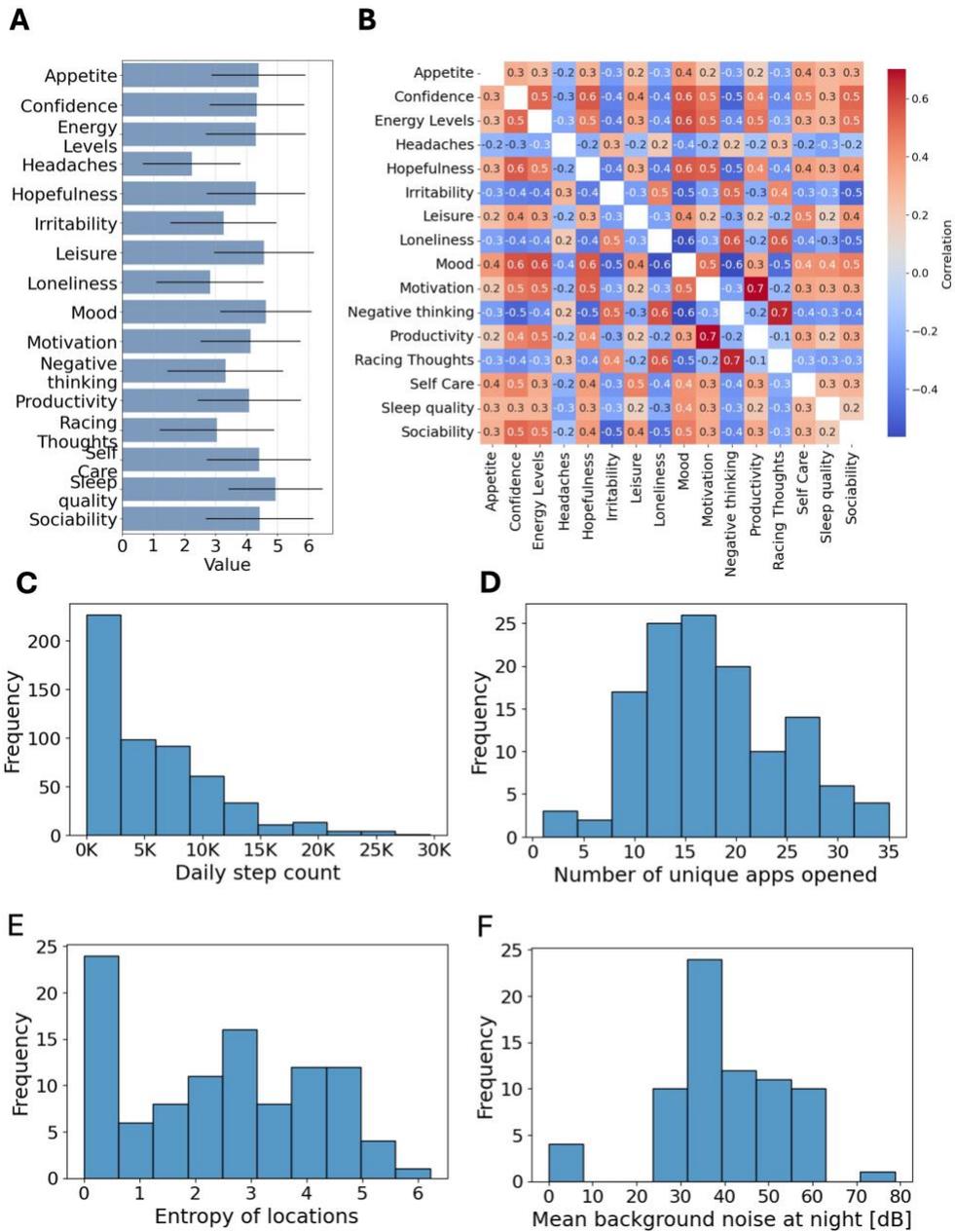

**Figure 3:** (A) Distribution of responses across active data features. (B) Correlation heatmap of active data features, showing relationships between mental health indicators. (C-F) Frequency distributions of passive data features: daily step count (C), number of unique apps opened per day (D), location entropy reflecting movement variability (E), and mean background noise levels at night (F).

**Performance of Models Predicting Mental Health Outcomes**

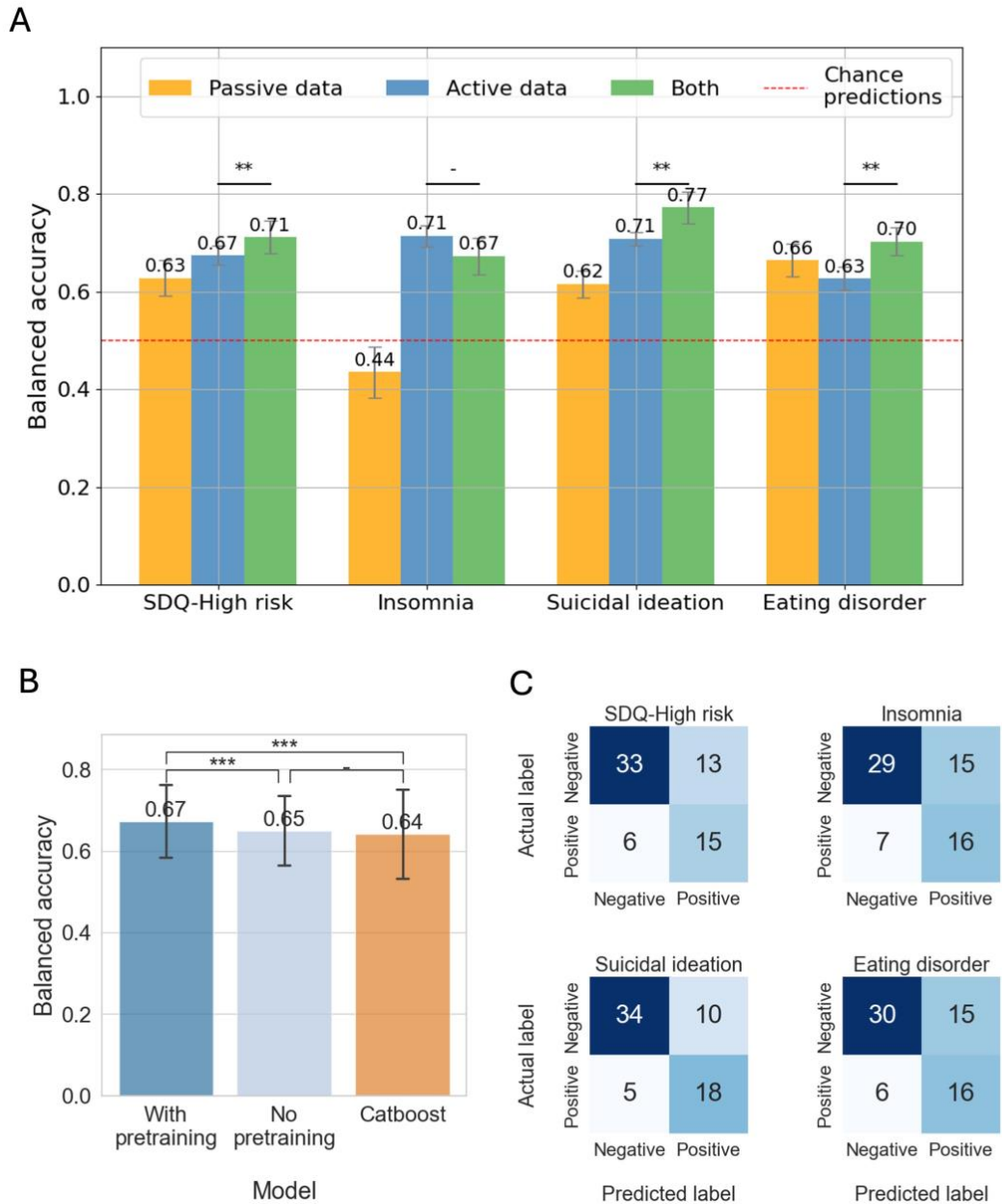

**Figure 4:** (A) Balanced accuracy of mental health outcome predictions (SDQ-high risk, insomnia, suicidal ideation, and eating disorder) using passive, active, and combined data. The red dashed line shows chance-level accuracy, and statistically significant differences are indicated (*: $P<.05$, **: $P<.01$, ***: $P<.001$, Wilcoxon signed-rank test). (B) Comparison of balanced accuracy for models with contrastive pretraining, without pretraining, and a CatBoost model, showing the performance benefit of pretraining ($P<.001$, paired t-test). (C) Confusion matrices for the combined data model's predictions, showing true positive and negative classifications across mental health outcomes.

Figure 4A illustrates the balanced accuracy of predictive models for mental health outcomes—SDQ-High risk, insomnia, suicidal ideation, and eating disorder—evaluated across 10 repetitions of leave-one-subject-out cross-validation. The analysis involved three feature sets: passive data, active data, and a combination of both. This evaluation was restricted to the 67 participants who provided both active and passive data to ensure fairness in comparison.

For SDQ-High risk, the model using passive data achieved a balanced accuracy of 0.63, while active data alone reached 0.67. The combined model, leveraging both data types, achieved a significantly higher balanced accuracy of 0.71 compared to active data alone ($P=.003$, Wilcoxon signed-rank test). Similarly, the combined data model outperformed the active data alone for eating disorder predictions, with balanced accuracies of 0.70 and 0.63, respectively ($P=.003$, Wilcoxon signed-rank test). In predicting insomnia, the combined model achieved a balanced accuracy of 0.67, while passive data alone performed below the chance level (0.44). For suicidal ideation, the combined model achieved the highest balanced accuracy of 0.77, significantly outperforming both active data (0.71) and passive data (0.62, $P=.003$, Wilcoxon signed-rank test). Table 3 summarises additional performance metrics, including AUC, AUC-PR, F1 scores, sensitivity, specificity, precision, and recall for each mental health outcome.

**Table 3: Detailed Performance Metrics for Mental Health Outcome Predictions**

|  | SDQ-High Risk | Insomnia | Suicidal ideation | Eating disorder |
|---|---|---|---|---|
| **Balanced accuracy** | 0.71 ± 0.03 | 0.67 ± 0.04 | 0.77 ± 0.03 | 0.70 ± 0.03 |
| **AUC** | 0.77 ± 0.03 | 0.74 ± 0.02 | 0.82 ± 0.03 | 0.73 ± 0.02 |
| **AUC-PR** | 0.53 ± 0.04 | 0.52 ± 0.05 | 0.64 ± 0.05 | 0.52 ± 0.03 |
| **F1** | 0.61 ± 0.04 | 0.59 ± 0.04 | 0.70 ± 0.04 | 0.61 ± 0.03 |
| **F1 macro** | 0.69 ± 0.03 | 0.66 ± 0.04 | 0.76 ± 0.03 | 0.68 ± 0.03 |
| **Sensitivity** | 0.71 ± 0.06 | 0.68 ± 0.03 | 0.78 ± 0.05 | 0.74 ± 0.03 |
| **Specificity** | 0.71 ± 0.05 | 0.66 ± 0.07 | 0.77 ± 0.04 | 0.67 ± 0.04 |
| **Precision** | 0.53 ± 0.04 | 0.52 ± 0.05 | 0.64 ± 0.05 | 0.52 ± 0.03 |
| **Recall** | 0.71 ± 0.06 | 0.68 ± 0.03 | 0.78 ± 0.05 | 0.74 ± 0.03 |

Figure 4B demonstrates the effectiveness of contrastive pretraining. Models with pretraining achieved the highest balanced accuracy (0.67), significantly outperforming both the model without pretraining (0.65, $P<.001$, paired t-test) and the CatBoost model (0.64, $P<.001$, paired t-test).

Figure 4C presents the confusion matrices for the combined data models. For SDQ-High risk, the model correctly identified 33 negatives and 15 positives, with six false negatives and 13 false positives. The model had higher misclassification rates for insomnia, with seven false negatives and 15 false positives. In predicting suicidal ideation, the model demonstrated strong performance, correctly classifying 34 negatives and 18 positives, with only five false negatives and 10 false positives. Similarly, for eating disorder, the model accurately identified 30 negatives and 16 positives, with six false negatives and 15 false positives.

We also examined (Supplementary Figure 2) whether predictive performance for active data alone differed between all study participants (N=103) and the subset who provided both active and passive data (N=67). The balanced accuracy remained consistent across both groups.

### Predictive Accuracy Across Mental Health Risk Groups

Figure 5 illustrates model accuracy in predicting mental health risks using combined active and passive data, segmented by risk levels for various mental health measures. The model performed exceptionally well at extreme risk levels, achieving near-perfect accuracy for high-risk groups (e.g., SCI scores 0–8, EDQ scores 4–6) and low-risk groups (e.g., SDQ scores 1–8). However, accuracy decreased significantly in ranges near thresholds (e.g., SDQ scores 9–16, SCI scores 9–16).

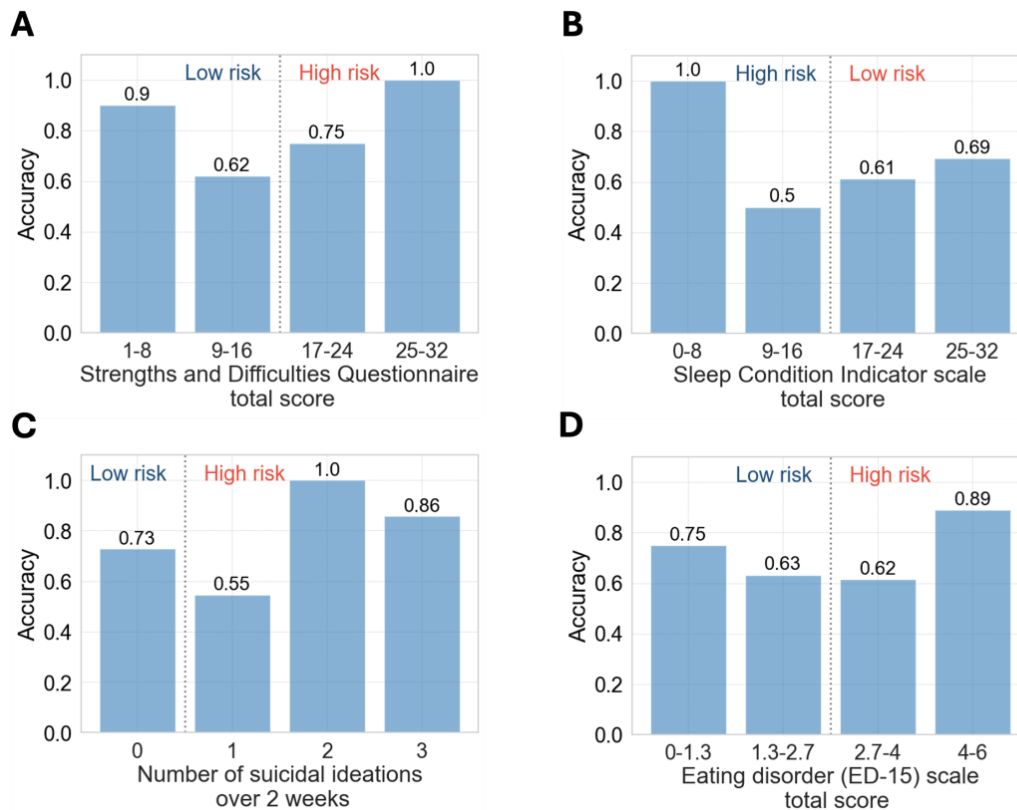

**Figure 5:** Accuracy of mental health risk prediction across different levels of (A) SDQ total score, (B) Sleep Condition Indicator (SCI) score, (C) Frequency of suicidal ideation thoughts, and (D) Eating Disorder (ED-15) total score.

## Model Interpretability: Active and Passive Data Contributions

Figure 6A illustrates the feature importance calculated using SHAP values for predicting the SDQ high-risk category using a combination of both active and passive data, with passive data aggregated by sensor type and active data shown individually. The top predictors included negative thinking, location features, app usage, racing thoughts, and self-care. Cognitive and emotional indicators (e.g., negative thinking, racing thoughts) ranked highest among active data features, while movement and environmental stability (e.g., location entropy, step count) dominated passive data contributions.

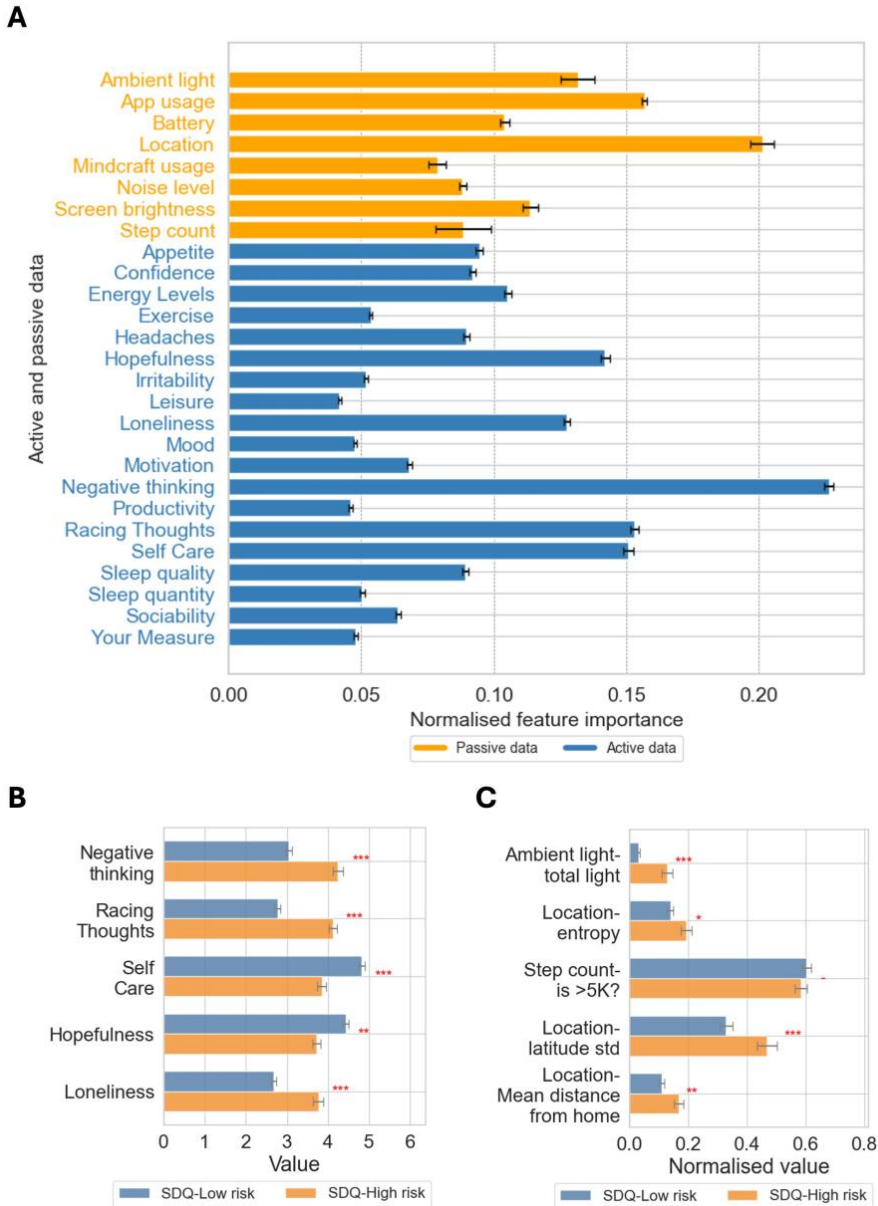

**Figure 6:** Feature importance analysis for predicting the SDQ high-risk category using both active and passive data. (A) SHAP-based feature importances, with passive data aggregated by sensor type and active data shown individually. (B) Distribution of the top five active data features across SDQ risk categories. (C) Distribution of the top five passive data features across SDQ risk categories. Statistically significant differences between low-risk and high-risk groups are indicated ((*: *P*<.05, **: *P*<.01, ***: *P*<.001, t-test).

The distribution of the top five active data features—negative thinking, racing thoughts, self-care, hopefulness, and loneliness—showed clear distinctions between low- and high-risk SDQ groups (Figure 6B). Negative thinking and racing thoughts were significantly higher in the high-risk group ($P<.001$, t-test). Conversely, self-care ($P<.001$, t-test) and hopefulness ($P<.001$, t-test) were significantly lower. Loneliness was also notably higher in the high-risk group ($P<.001$, t-test).

The distribution of the top five passive data features—ambient light, location entropy, step count, latitude standard deviation, and mean distance from home—highlighted significant differences between risk groups (Figure 6C). High-risk individuals showed greater ambient light exposure ($P<.001$, t-test), potentially reflecting greater exposure to light at night and sleep disruptions. They also exhibited higher location entropy ($P=.016$, t-test) and latitude variability ($P<.001$, t-test). Additionally, fewer high-risk individuals exceeded 5,000 daily steps ($P=.002$, t-test).

## Discussion

### Principal findings

Our study demonstrated the effectiveness of integrating active self-reported and passive smartphone sensor data to predict adolescent mental health risks using a novel machine learning framework. By leveraging data collected via the Mindcraft app, we evaluated predictions across four critical mental health outcomes: SDQ-High risk, insomnia, suicidal ideation, and eating disorders. Combined models consistently outperformed those based on active or passive data alone, with balanced accuracies of 0.71 for SDQ-High risk, 0.77 for suicidal ideation and 0.70 for eating disorders. These results highlight the complementary value of passive data, which unobtrusively captures continuous behavioural patterns that enrich the subjective insights provided by active data.

The user engagement patterns over the 14-day study period underscore the sustained utility of passive data collection in longitudinal studies, given its lower participant burden. Active data engagement declined more rapidly than passive data, highlighting the feasibility of using unobtrusive, passive metrics in scalable mental health monitoring frameworks. Additionally, users preferred less intrusive metrics such as step count, battery usage, and screen brightness, suggesting the importance of prioritising user-friendly data collection methods.

Passive data features such as location entropy, ambient light levels, and step count were identified as clinically relevant predictors for the high-risk SDQ group, reflecting behavioural and lifestyle indicators of mental health risks. Fewer high-risk individuals exceeded 5,000 daily steps, reinforcing the established link between physical inactivity and mental health risk. Location entropy, which measures the unpredictability or variability in an individual's movement patterns, may suggest a lack of routine or stability in daily life, potentially indicative of less structured or more uncertain environments. This finding, while tentative, highlights the potential of digital phenotyping to reveal novel behavioural markers of mental health risk, warranting further exploration in future studies to validate its significance.

Similarly, active data features like negative thinking, racing thoughts, and self-care provided critical insights into internalising and externalising behaviours, with significant distinctions between high- and low-risk groups. Notably, these features appear to aggregate symptoms of both internalising and externalising behaviours; for example, racing thoughts likely capture aspects of worry, indicative of internalising symptoms, as well as cognitive hyperactivity or distractibility, characteristic of externalising symptoms.

Importantly, our innovative contrastive learning approach proved effective in addressing the variability inherent in daily behavioural data. The pretraining phase enhanced the robustness of model predictions by stabilising user-specific feature representations. This methodological advancement yielded improved balanced accuracy and increased confidence in the model's applicability to the real world.

### Comparison with prior work

Our study makes important contributions to the evolving field of digital phenotyping for mental health. Previous studies leveraging passive sensing to support mental health and well-being [15, 21-31] have largely focused on adults, and research on CYP remains scarce [33, 49].

While several studies [22, 30-33] have investigated the use of passive sensing for adolescent mental health, they often face limitations in scope and methodology. Previous research [22, 30-32] has primarily focused on specific mental health outcomes, such as depression and anxiety, in clinical populations, limiting the generalisability of their findings to broader, non-clinical groups. Digital self-monitoring has a potential role in multiple stages of the clinical pathway, from prevention to clinical intervention. Our work addresses a broader range of mental health outcomes— internalising and externalising disorders, eating disorders, insomnia and the presence of suicidal ideation, in a non-clinical, non-help-seeking adolescent population. MacLeod et al. [33], the closest study to ours, included adolescents from clinical and non-clinical settings but relied solely on passive sensing, with no participant interaction beyond the initial setup. To the best of our knowledge, this study is the first to use ML to accurately predict mental health risk across a broad range of outcomes in low and higher-risk young people using a combination of active and passive data in a general adolescent population.

### Strength and limitations
This study has several notable strengths that contribute to advancing adolescent mental health prediction. By integrating active self-reports and passive sensor data through the accessible Mindcraft app, the study offers a scalable and practical approach that bridges subjective and objective measures. Notably, our models performed well despite high attrition in active data, addressing a common criticism in the field that participants often stop tracking after a few days. This highlights the robustness of passive data collection and its potential for scalable, long-term applications, such as early detection of mental health problems through school-based screening.

An innovative methodological strength is the use of contrastive learning to stabilise user-specific feature representations. This approach mitigates the variability in daily behavioural data, improving model robustness and generalisability. The study's leave-one-subject-out cross-validation framework highlights its reliability in capturing inter-subject variability, a critical factor for real-world applicability in digital phenotyping.  Furthermore, the incorporation of SHAP-based feature interpretability enhances transparency and clinical relevance by identifying key predictors. This makes the model more understandable and thus more likely to be adopted by clinicians and young people, addressing common criticisms of 'AI black box' approaches that lack explainability [50]. By fostering trust, this transparency supports the broader adoption of digital mental health tools in clinical and community settings.

Despite these strengths, the study has several limitations that warrant consideration. While sufficient for feasibility testing, the sample size is relatively small and restricted to London schools, which may affect the generalisability of the findings to broader populations. Moreover, the gender distribution is skewed, with a higher proportion of female participants due to the inclusion of a girls-only school. Another limitation is the short two-week data collection period, which may not adequately capture the long-term fluctuations typical of mental health conditions. Future studies should aim for a more balanced and diverse sample and an extended data collection period to validate these findings across various demographic groups. Our model performance also needs improvement in identifying individuals with borderline SDQ scores who might develop mental disorders in the future and for whom digital phenotyping might be particularly helpful in informing early mental health intervention.

### Implications and Recommendations

Real-time digital phenotyping at a population level can complement traditional screening methods by identifying and prioritising high-risk individuals and tailoring intervention prevention and early intervention strategies [51, 52]. The use of a mobile app for digital phenotyping is particularly vital for CYP, for whom early identification and intervention are essential to prevent the onset of more severe mental health issues in adulthood. When implemented in schools, it addresses barriers such as stigma and accessibility, offering adolescents a preventive tool that empowers them to manage their mental health. Digital phenotyping also offers the opportunity to inform school-based digital interventions that might be central to the early intervention and prevention of mental health problems in the community [53].

To maximise utility, digital tools like Mindcraft should be integrated into population health strategies. For example, mental health in schools or primary care healthcare providers could use app-generated insights to monitor progress between assessments, bridging gaps in care and facilitating personalised interventions comparable to their utilisation in clinical care [51, 54] . By fostering trust and widespread adoption, digital phenotyping can bridge critical gaps in care and transform the future of personalised healthcare [55, 56], including the potential for seamless integration into clinical workflows, subject to robust privacy safeguards such as secure data storage and transparent consent mechanisms. With their ubiquity and affordability, smartphones enable continuous, real-time data collection, even in low-resource settings, thus reducing reliance on clinical oversight [31, 57].

Current platforms such as Childline rely on proactive engagement from children and young people, creating barriers for disengaged users. In contrast, Mindcraft's passive tracking capabilities offer a proactive approach by identifying early signs of poor mental health or well-being and prompting timely professional interventions. With further development, Mindcraft could evolve into a comprehensive platform, integrating in-app support (e.g. behavioural recommendations and counselling services) informed by active and passive data. This integration of proactive detection and tailored intervention has the potential to address significant gaps in traditional mental health support systems.

Despite the growing interest in recommendation systems within healthcare research [58-62], their application to mental health remains limited. Mindcraft presents an opportunity to integrate personalised recommendation systems that leverage user profiles to offer tailored mental health interventions. Future iterations could incorporate reinforcement learning algorithms to adaptively suggest mood-enhancing activities such as mindfulness exercises, physical activities, or cognitive techniques to improve mental health [63, 64]. This personalised approach has the potential to improve user engagement and the effectiveness of interventions.

## Conclusion

In conclusion, this study underscores the transformative potential of integrating active and passive smartphone data for adolescent mental health prediction. By leveraging innovative ML techniques, such as contrastive learning, and the scalability of tools like the Mindcraft app, we present a robust framework for early risk detection across diverse mental health outcomes. These findings lay the groundwork for more inclusive, accessible, and personalised early detection and intervention strategies in adolescent mental health.

## Author Contributions

DN, MDS, and AAF conceptualised the study. TBB, DN and MDS gained ethical approval and conducted recruitment and data collection. BK curated the data and performed data analysis and machine learning under the supervision of AAF. BK, TBB, and AF drafted the original manuscript. MDS, DN and AAF critically reviewed and edited the manuscript. All authors reviewed and approved the final version of the manuscript for publication.

## Acknowledgements
TBB is supported by a Fellowship funded by the Fundación Alicia Koplowitz. DN is supported by the National Institute for Health Research (NIHR) Applied Research Collaboration Northwest London and NIHR Imperial Biomedical Research Collaboration. MDS is supported by the NIHR Imperial Biomedical Research Collaboration and NIHR Mental Health Translational Research Collaboration. AAF acknowledges support from the United Kingdom Research and Innovation Turing AI Fellowship (EP/V025449/1).

## Data Availability
The study data are not publicly available because of the General Data Protection Regulation restrictions and privacy policies outlined in the participant recruitment. The data sets generated during this study are available from the corresponding author upon reasonable request.

## Use of Generative AI
We acknowledge using Grammarly to identify grammatical errors and improvements to the writing style.

## Conflicts of Interest
None declared

## Abbreviations
CYP: Children and Young People
ED: Eating disorder
ML: Machine learning
SDQ: Strengths and Difficulties Questionnaire

**Supplementary Figures**

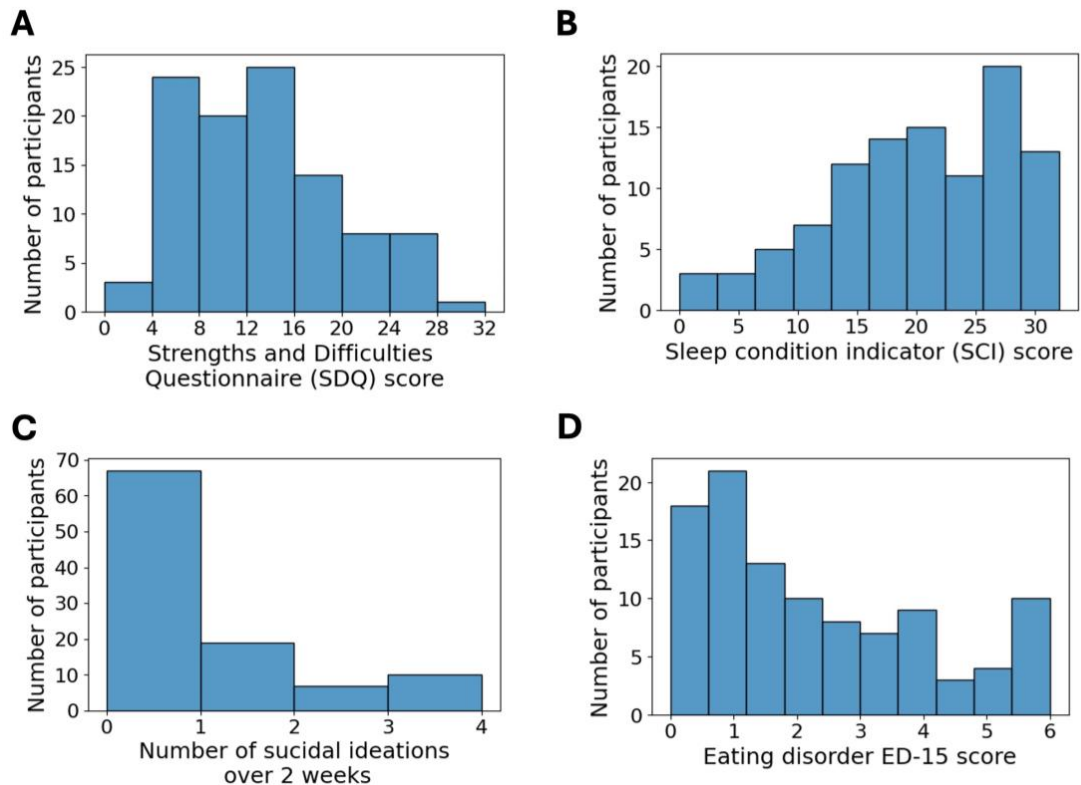

**Supplementary Figure 1:** Distribution of mental health assessment scores across participants. (A) Strengths and Difficulties Questionnaire (SDQ) scores, representing overall mental health and behavioural difficulties. (B) Sleep Condition Indicator (SCI) scores, assessing sleep quality and potential insomnia. (C) Frequency of self-reported suicidal ideations over a two-week period. (D) Eating Disorder Examination Questionnaire (ED-15) scores, evaluating symptoms associated with eating disorders.

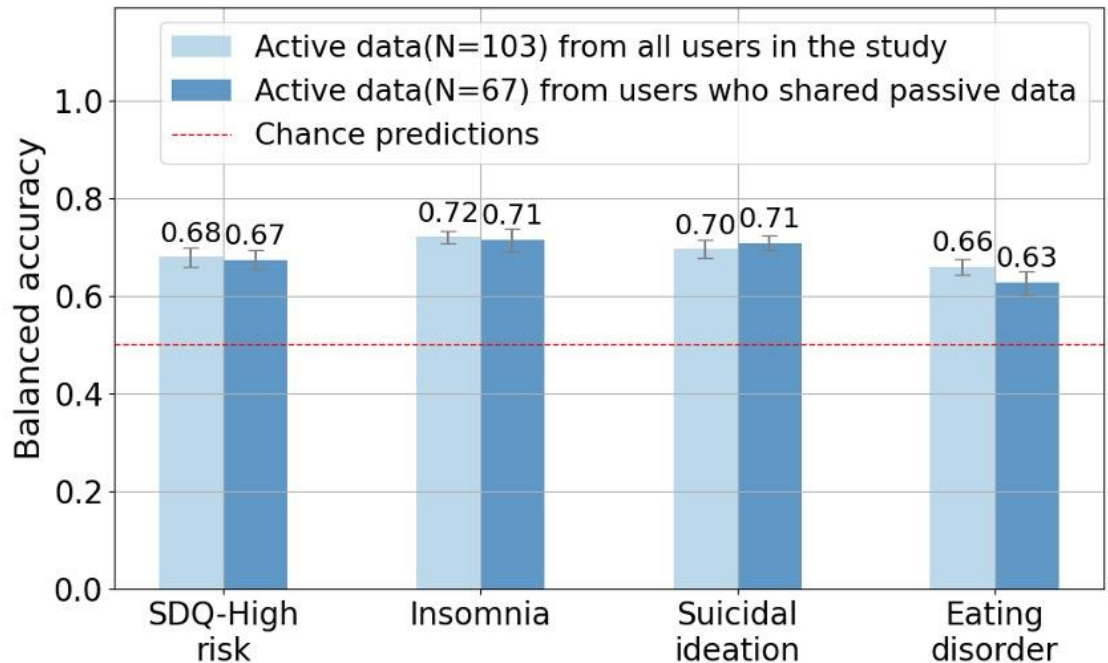

**Supplementary Figure 2:** Comparison of balanced accuracy of mental health outcome predictions using active data only for all users in the study (N=103) vs users who also enabled passive data collection (N=67).